\pdfoutput=1

\documentclass[11pt]{article}

\usepackage[]{acl}

\usepackage{times}
\usepackage{latexsym}

\usepackage[T1]{fontenc}

\usepackage[utf8]{inputenc}

\usepackage{microtype}

\usepackage{inconsolata}

\usepackage{tabularx}
\usepackage{booktabs}
\usepackage{multirow}
\usepackage{graphicx}
\usepackage{tablefootnote}

\title{Fine-grained Gender Control in Machine Translation with Large Language Models}

\author{Minwoo Lee\textsuperscript{1,2}\hspace{1cm} Hyukhun Koh\textsuperscript{3} \hspace{1cm} {\bf Minsung Kim\textsuperscript{2}} \hspace{1cm} {\bf Kyomin Jung\textsuperscript{2,3,4}}\\
  \textsuperscript{1}LG AI Research, \hspace{5mm}
  \textsuperscript{2}Dept. of ECE, Seoul National University, \\
  \textsuperscript{3}IPAI, Seoul National University, \hspace{5mm}
  \textsuperscript{4}Institute of Engineering Research, Seoul National University \\
  \texttt{minwoo.lee@lgresearch.ai} \hspace{5mm}
  \texttt{\{hyukhunkoh-ai, kms0805, kjung\}@snu.ac.kr} \\}

\begin{document}
\maketitle

\begin{abstract}

In machine translation, the problem of ambiguously gendered input has been pointed out, where the gender of an entity is not available in the source sentence.
To address this ambiguity issue, the task of controlled translation that takes the gender of the ambiguous entity as additional input have been proposed. 
However, most existing works have only considered a simplified setup of one target gender for input.
In this paper, we tackle controlled translation in a more realistic setting of inputs with multiple entities and propose \textbf{Gender-of-Entity (GoE) prompting} method for LLMs.
Our proposed method instructs the model with fine-grained entity-level gender information to translate with correct gender inflections.
By utilizing four evaluation benchmarks, we investigate the controlled translation capability of LLMs in multiple dimensions and find that LLMs reach state-of-the-art performance in controlled translation. 
Furthermore, we discover an emergence of \textit{gender interference} phenomenon when controlling the gender of multiple entities.
Finally, we address the limitations of existing gender accuracy evaluation metrics and propose leveraging LLMs as an evaluator for gender inflection in machine translation.\footnote{Code available at \url{https://github.com/minwhoo/fine-grained-gender-control-mt}}

\end{abstract}

\section{Introduction}
In machine translation (MT) research, many efforts have been made to improve the gender accuracy of NMT systems, which have shown to exhibit gender bias \citep{savoldi-etal-2021-gender,Piazzolla2023GoodBN}. 
This research includes the task of handling ambiguously gendered entities in text, which arises from differences in gender markings across different languages \citep{Bentivogli2020GenderID}. 
Without consideration for these ambiguities, existing MT systems default to masculine translations or use a stereotypically associated gender, reflecting the bias in training data \citep{Cho2019OnMG}.

To address the gender ambiguity issue, multiple approaches have been proposed, such as gendered translation rewriting \citep{Rarrick2023GATEAC}, generating gender-neutral translations \citep{Piergentili2023HiGO}, and controlled translation \citep{Bentivogli2020GenderID}. 
Specifically, controlled translation methods take the gender of the ambiguous entity as additional input along with the source text, and generate a translation matching the given gender. 
However, most previous works have only considered gender control of a single entity for each input, and a gap still remains between this simplified experimental setup and texts found in real-world contexts where multiple entities are often mentioned within the same context.

In this work, we investigate the task of controlled translation in a more realistic, \textit{fine-grained} setting where the given text has multiple entities with different gender assignments. 
To this end, we employ LLMs and propose \textbf{Gender-of-Entity (GoE) prompting} for fine-grained gender control in machine translation.
Our method utilizes the powerful instruction-following and translation capabilties of LLMs for a more accurate translation aligned with the target gender inflections. 
LLMs are explicitly instructed to translate the source text with additional entity-level gender information given in natural language statements.

For a comprehensive assessment, we employ four existing benchmarks on gender bias evaluation and investigate the LLM's capabilities in various scenarios, ranging from sentences with multiple ambiguously gendered entities to sentences containing both unambiguously gendered and ambiguously gendered entities. 
From our experiments, we find that the GoE prompting on the LLMs scores up to an average of 95.4\% gender accuracy on the Must-SHE dataset, significantly outperforming previous control methods based on fine-tuning. 
Furthermore, we identify a problematic phenomenon of \textit{gender interference} in fine-grained controlled translation, where controlling the gender of one entity adversely affects the gender inflection of other entities.
These findings emphasize the necessity of fine-grained assessment of gendered entities in gender bias evaluation.

Finally, we find that conventional metrics used in gender bias evaluation are based on lexical matching, making it challenging to capture synonyms or paraphrases. 
We thus propose leveraging LLMs as a reference-free evaluator that checks the gender inflections and agreements of the translation. 
Experimental results show the validity of our proposed evaluator from the high correlation with human judgements and with the automated metrics as well. 

To summarize our work, in Section 2, we formulate the fine-grained controlled machine translation task and introduce the four benchmark datasets used in our paper. Next, we report our controlled translation experiments on the four evaluation settings and share our findings in Section 3. In Section 4, we investigate using LLMs as a gender evaluator. In Section 5, we share related works and conclude our paper in Section 6.

\section{Gender Control in Machine Translation}
We formalize the controlled translation task of gender attributes in machine translation and introduce four gender control scenarios based on existing evaluation benchmarks. 
We then introduce our proposed LLM-based controlled translation methodology. 

\subsection{Task definition}

In our study, we consider the controlled translation task where one or more entities in the source text are directed to have a specified gender inflection in the target translation output. 
We approach the task in a \textit{fine-grained} setting, where we control the gender inflection of each entity in the text separately. 

We formalize the controlled translation task as follows: 
given a source sentence $src_i$ and a mapping that assigns a specific gender to each entity in $src_i$, produce a target translation with the correct gender inflections matching the given mapping. 
We will refer to the set of controlled entities $E_i$, set of target genders $\mathcal{G}$, entity-gender mapping $f_i: E_i \rightarrow \mathcal{G}$, and the target translation with matching gender inflections $tgt_i^{f_i}$:
$$controlled\_translation(src_i, f_i) \rightarrow tgt_i^{f_i}.$$

In our study, we limit the set of target genders $\mathcal{G}$ to masculine and feminine supported by the evaluation datasets. 

\subsection{Evaluation benchmarks}

\begin{figure}[t]
\centering{
\includegraphics[width=\columnwidth]{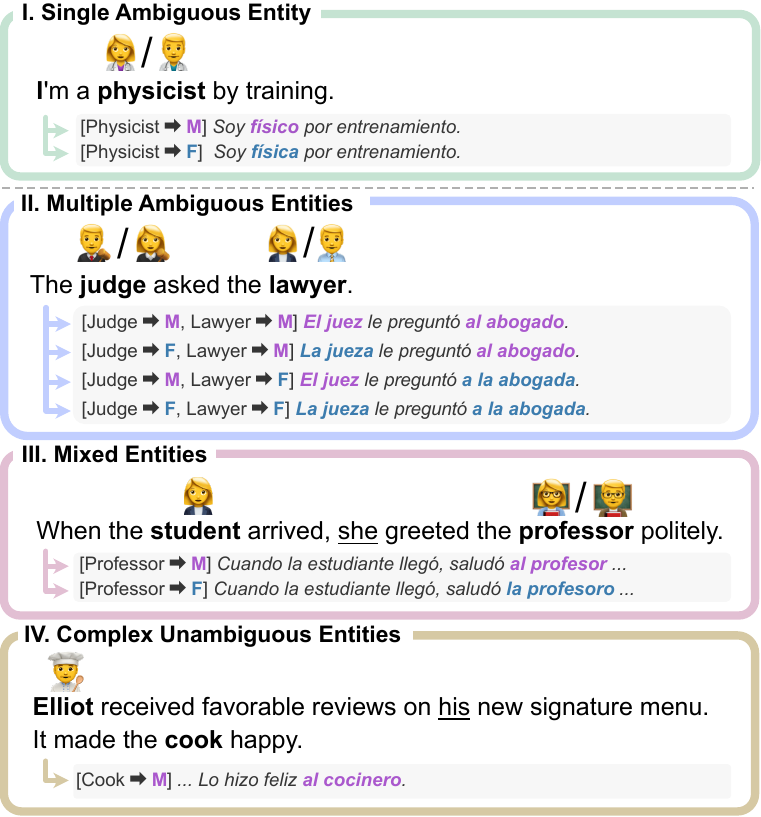}
}
\vspace{-5mm}
\caption{Four gender control scenarios in machine translation investigated in our study. 
}
\label{fig:task_categories}
\vspace{-5mm}
\end{figure}

In order for a comprehensive assessment of fine-grained controlled translation, we employ four existing gender evaluation datasets in our work. 
The evaluation benchmarks have been constructed with different objectives,
enabling multi-faceted analysis of controlled translation, which we categorize into four scenarios, as shown in Figure~\ref{fig:task_categories}. 

\textbf{Single Ambiguous Entity} We first evaluate the controlled translation of sentences with a single ambiguously gendered entity using the MuST-SHE dataset \citep{Bentivogli2020GenderID}, constructed from parallel transcripts from TED talks. We specifically use the \textit{Category 1} subset, which consists of sentence pairs that require knowledge of the speaker's gender for the correct translation. 

\textbf{Multiple Ambiguous Entities} Next, we evaluate the controlled translation of sentences with multiple ambiguous entities via the recently released GATE dataset \citep{Rarrick2023GATEAC}. The dataset consists of linguistically diverse sentences with multiple alternative target language translations constructed with the help of bilingual linguists.

\textbf{Mixed Entities} Thirdly, we evaluate controlled translation of sentences where both ambiguously gendered entities and unambiguously gendered entities co-exist via the widely used WinoMT benchmark \citep{stanovsky-etal-2019-evaluating}. The dataset consists of synthetically constructed sentences containing exactly two entities, of which only one is unambiguously gendered. While most works that utilize the dataset usually consider only the unambiguous entity,
we adopt the extension for evaluating the ambiguous entity by \citet{saunders-etal-2020-neural}.

\textbf{Complex Unambiguous Entities} Finally, we evaluate controlled translation where the entity is unambiguously gendered but hard to disambiguate due to the complex structure of the source text. For this scenario, we employ the \textit{Contextual} subset in the MT-GenEval dataset \citep{currey-etal-2022-mt}. The samples in this subset consist of two sentences, where the gender of the entity in the second sentence can only be inferred via the first sentence, as illustrated in Figure~\ref{fig:task_categories}.

For evaluation, we experiment controlled translation in three language directions supported by all four benchmarks: English to Spanish, English to French, and English to Italian. 
For dataset statistics and preprocessing details, refer to Appendix~\ref{sec:appendix}.

\subsection{Evaluation metrics}

\begin{table}[t]
    \centering
    \small
\begin{tabularx}{\columnwidth}{X}
\toprule
\textbf{Gender-of-Entity Prompting Template} \\
\midrule
SYSTEM: You are a professional \texttt{[TGT\_LANG]} translator that especially considers translating gender inflections correctly. \\
\\
USER: Translate the following sentence into \texttt{[TGT\_LANG]} (\texttt{[GENDER\_ANNOTATION]}): \texttt{[SRC]} \\
\midrule
\textbf{Gender Annotation} \\
\midrule
for \texttt{[ENT\_1]}, use \texttt{[GENDER\_1]}; ...; for \texttt{[ENT\_n]}, use \texttt{[GENDER\_n]} \\
\bottomrule
\end{tabularx}
\caption{Instruction template for our proposed Gender-of-Entity (GoE) prompting.}
\vspace{-5mm}
\label{tab:prompt}
\end{table}

We use the term-level coverage and accuracy defined by \citet{Bentivogli2020GenderID} for evaluating gender accuracy on all benchmarks, excluding WinoMT, which does not have the target gender annotations required for this metric.
\textbf{Coverage} is defined by the proportion of (either correct or incorrect) gendered terms that are lexically matched in the generated translation. 
\textbf{Accuracy} is subsequently defined by the proportion of correct terms out of all covered terms in the corpus.

Alternatively, the gender accuracy metric defined by \citet{stanovsky-etal-2019-evaluating} is used for the WinoMT dataset. The metric is based on a source-target alignment-based algorithm used jointly with a language-specific gender morphology analyzer to check if the gendered terms are correctly inflected.

For evaluating translation quality, we utilize the BLEU score, an n-gram based lexical metric, and COMET score \citep{rei-etal-2022-comet}, a neural metric that has been shown to be closely aligned with human judgments. 
\footnote{The sacrebleu id for computing bleu is: \texttt{s:1000|rs:12345|c:mixed|e:no|tok:13a|s:exp|v:2.3.1} and for COMET, we use the \texttt{Unbabel/wmt22-comet-da}.}

\subsection{Gender-of-Entity prompting for LLMs}

We propose \textbf{Gender-of-Entity (GoE) prompting} in our work for fine-grained controlled translation of gender using LLMs. 
Our zero-shot approach builds upon LLM's translation and instruction-following capabilities to direct the LLM to translate with the specified gender for each entity.

The template for Gender-of-Entity prompting is shown in Table~\ref{tab:prompt}, where \texttt{[TGT\_LANG]} is the slot for the name of the target translation language, \texttt{[SRC]} is the slot for the source text, and \texttt{[GENDER\_ANNOTATION]} is the slot where we specify the entity-level gender mappings in natural language. 
By default, we use an entity-level gender annotation scheme where we list the entities and their target gender, delimited by ``;''. 
More specifically, \texttt{[ENT\_i]} is substituted with the entity name found in source text, and \texttt{[GENDER\_i]} is substituted by either ``\textit{he/him}'' or ``\textit{she/her}'' for male and female gender inflections respectively.

We use two instruction-tuned LLMs, Llama 2 70B Chat and ChatGPT 3.5 (\texttt{gpt-3.5-turbo}) for applying GoE prompting to LLMs. The two models have shown to have competitive translation performance for the three language directions evaluated in our study \citep{zhu2023multilingual}.

\section{Main Experiments}
\label{Main Experiments}
We experiment on controlled translation of gender with our proposed method on the four evaluation benchmarks and compare them with existing approaches. 

\subsection{Gender Control of Single Ambiguous Entity}

\begin{table*}[h]
    \centering
    \small
\setlength\tabcolsep{4.5pt}
\begin{tabularx}{\textwidth}{X|cccc|cccc|cccc}
\toprule
 & \multicolumn{4}{c}{\textbf{ES}} & \multicolumn{4}{c}{\textbf{FR}} & \multicolumn{4}{c}{\textbf{IT}} \\
 
\textbf{Method} & Cov. & Acc. & \textsc{bleu} & \textsc{comet} & Cov. & Acc. & \textsc{bleu} & \textsc{comet} & Cov. & Acc. & \textsc{bleu} & \textsc{comet} \\
\midrule
\textit{NLLB-200 600M D.} & & & & & & & & & & & & \\

Baseline      & \color{gray} 74.6 & \color{gray} 53.9 & \color{gray} 43.7 & \color{gray} 85.1 & \color{gray} 62.7 & \color{gray} 53.6 & \color{gray} 37.0 & \color{gray} 82.8 & \color{gray} 60.0 & \color{gray} 52.3 & \color{gray} 35.4 & \color{gray} 84.8 \\
Gender prefixing & 75.3 & 77.6 & 44.9 & 85.6 & 62.2 & 72.2 & 38.3 & 83.3 & 60.3 & 74.2 & 36.2 & 84.9 \\
CG* \scriptsize \citep{Liu2023HowTA}           & - & 82.8 & 44.7 & 84.7 & - & 79.4 & 38.7 & 82.5 & - & 83.6 & 35.4 &  83.7 \\
FT* \scriptsize  \citep{Liu2023HowTA}            & - & 86.9 & 43.7 & 84.0 & - & 85.0 & 38.2 & 82.0 & - & 87.8 & 34.4 & 83.5 \\
\midrule
\textit{NLLB-200 1.3B D.} & & & & & & & & & & & & \\
Baseline & \color{gray} 76.1 & \color{gray} 60.0 & \color{gray} 45.5 & \color{gray} 85.8 & \color{gray} 63.3 & \color{gray} 58.7 & \color{gray} 39.4 & \color{gray} 83.8 & \color{gray} 64.1 & \color{gray} 59.6 & \color{gray} 37.5 & \color{gray} 86.0 \\
Gender prefixing & 76.8 & 84.3 & \textbf{47.3} & 86.0 & 61.8 & 81.2 & \textbf{40.5} & 83.5 & 63.9 & 84.7 & \textbf{38.1} & 85.9 \\
\midrule
\textit{Llama 2 70B Chat} & & & & & & & & & & & & \\
Baseline & \color{gray} 69.0 & \color{gray} 54.8 & \color{gray} 34.4 & \color{gray} 82.4 & \color{gray} 53.4 & \color{gray} 54.3 & \color{gray} 29.5 & \color{gray} 80.2 & \color{gray} 54.1 & \color{gray} 54.1 & \color{gray} 28.5 & \color{gray} 81.9 \\
GoE prompting & 71.0 & 94.9 & 37.6 & 83.5 & 57.6 & 94.0 & 31.9 & 81.6 & 54.5 & 89.8 & 30.0 & 82.4 \\
\midrule
\textit{ChatGPT 3.5} & & & & & & & & & & & & \\
Baseline & \color{gray} 73.4 & \color{gray} 54.0 & \color{gray} 39.1 & \color{gray} 85.5 &  \color{gray} 42.0 & \color{gray} 56.7 & \color{gray} 33.1 & \color{gray} 83.5 & \color{gray} 63.1 & \color{gray} 51.9 & \color{gray} 33.5 & \color{gray} 86.1 \\
GoE prompting & 77.1 & \textbf{96.5} & 42.7 & \textbf{87.0} & 64.9 & \textbf{95.3} & 37.9 & \textbf{85.3} & 63.9 & \textbf{94.4} & 35.4 & \textbf{86.7} \\
\bottomrule
\end{tabularx}
\caption{Results of controlled translation on the Must-SHE dataset. Gray text denote baseline results without the gender specified. *Results are taken from \citet{Liu2023HowTA} }
\vspace{-5mm}
\label{tab:mustshe}
\end{table*}

First, we consider the most straightforward setup where there is a single ambiguously gendered entity in the source sentence. 
We evaluate on the MuST-SHE benchmark \citep{Bentivogli2020GenderID}, where we control the ambiguous entity to the designated gender label provided by the annotation. 
Since the ambiguously gendered entity is always the speaker for this dataset, we use the gender annotation ``\textit{the speaker is male}'' or ``\textit{the speaker is female}'' depending on the designated gender.

\subsubsection{Baseline methods}

We compare our approach with three baseline methods developed for pre-trained NMT models: gender prefixing, gender-specific fine-tuning  (FT), and inference-time classifier guidance (CG) \citep{Liu2023HowTA}. 
Gender prefixing simply adds gendered prefixes ``\texttt{MALE:}'' and ``\texttt{FEMALE:}'' in front of the source text.
Gender-specific fine-tuning (FT) fine-tunes separate NMT models on a gendered parallel corpus for each gender. Finally, inference-time classifier guidance (CG) utilizes a pre-trained gender attribute classifier module to modify the decoder activations of existing NMT models during inference.
For gender prefixing, we share evaluation results on both NLLB-200 600M distilled and NLLB-200 1.3B distilled models, which are multilingual NMT models shown to have strong translation performance \citep{nllbteam2022language}.
For the fine-tuning and classifier-guidance approaches, we report results by \citet{Liu2023HowTA} on the NLLB-200 600M distilled model. 

\subsubsection{Results}

Experimental results, shown in Table~\ref{tab:mustshe}, indicate that GoE prompting is highly effective at controlling the gender of a single entity, reaching very high gender accuracies on both Llama 2 and ChatGPT 3.5 models and for all three target languages. Especially for ChatGPT, the accuracies are in the range of 94\% and 96\%, reaching state-of-the-art performance.
Furthermore, even though the baseline gender accuracy of NLLB-200 600M distilled model and LLMs have similar scores, the improvement from our zero-shot prompting exceeds the improvement from existing baseline approaches that require fine-tuning. This highlights the strong zero-shot instruction following capabilities of LLMs. 

In terms of translation quality, NLLB-200 models have the highest BLEU scores, followed by ChatGPT and Llama 2 models. 
Based on the COMET scores, however, ChatGPT scores the highest, followed by NLLB-200 models and Llama 2. 
These findings suggest that ChatGPT 3.5 LLMs have competitive zero-shot translation performance compared to the evaluated NLLB-200 models, while Llama 2 trails behind slightly. 

\subsection{Gender Control of Multiple Ambiguous Entities}
\label{Gender Control of Multiple Ambiguous Entities}
Next, we evaluate controlled translation on the GATE benchmark \citep{Rarrick2023GATEAC}, which consists of sentences with up to three ambiguously gendered entities.
The dataset also includes translations and annotations of all possible combinations of male/female gender mappings for each entity. This means a sentence with $N$ ambiguous entities will have  $2^N$ possible gender mappings and an equal number of translations.
We evaluate Llama 2 and ChatGPT on controlled translation to all possible gender mappings using the default GoE prompting template described in Table~\ref{tab:prompt}.

\subsubsection{Results}

\begin{table}[h]
    \centering
    \scriptsize
\setlength\tabcolsep{4.5pt}

\begin{tabularx}{\columnwidth}{llXr|rr|rr}
\toprule
 & & & & \multicolumn{2}{c}{\textbf{Gender}} & \multicolumn{2}{c}{\textbf{\#Ent}} \\
\textbf{Lang.} & \textbf{Model} & \textbf{Method} & Cov. & \multicolumn{1}{c}{Acc\textsubscript{M}} & \multicolumn{1}{c}{Acc\textsubscript{F}} & \multicolumn{1}{c}{Acc\textsubscript{1}} & \multicolumn{1}{c}{Acc\textsubscript{$\geq$2}} \\
\midrule
\multirow{4}{*}{ES} & \multirow{2}{*}{Llama 2} & Baseline & \color{gray}57.4 & \color{gray}88.7 & \color{gray}11.3 & \color{gray}50.0 & \color{gray}50.0  \\
                    &                          &      GoE & 62.2 & 97.9          & 68.1          & 84.9          & 81.3          \\
                    & \multirow{2}{*}{ChatGPT} & Baseline & \color{gray}66.5 & \color{gray}88.9 & \color{gray}11.1 & \color{gray}50.0 & \color{gray}50.0  \\
                    &                          &      GoE & 67.0 & \textbf{98.8} & \textbf{92.3} & \textbf{96.6} & \textbf{94.6} \\
\midrule
\multirow{4}{*}{FR} & \multirow{2}{*}{Llama 2} & Baseline & \color{gray}65.3 & \color{gray}95.4 & \color{gray}4.6  & \color{gray}50.0 & \color{gray}50.0 \\
                    &                          & GoE      & 66.2 & \textbf{97.5} & 58.8          & 82.7          & 74.1          \\
                    & \multirow{2}{*}{ChatGPT} & Baseline & \color{gray}71.7 & \color{gray}88.9 & \color{gray}11.1 & \color{gray}50.0 & \color{gray}50.0  \\
                    &                          & GoE      & 69.9 & 96.4          & \textbf{81.0} & \textbf{91.3} & \textbf{86.4} \\
\midrule
\multirow{4}{*}{IT} & \multirow{2}{*}{Llama 2} & Baseline & \color{gray}62.1 & \color{gray}94.4 & \color{gray}5.6 & \color{gray}50.0 & \color{gray}50.0 \\
                    &                          & GoE      & 61.2 & \textbf{98.7} & 49.7          & 71.3          & 75.6          \\
                    & \multirow{2}{*}{ChatGPT} & Baseline & \color{gray}72.6 & \color{gray}94.8 & \color{gray}5.2 & \color{gray}50.0 & \color{gray}50.0 \\
                    &                          & GoE      & 71.8 & 98.2          & \textbf{77.9} & \textbf{89.8} & \textbf{87.3} \\
\bottomrule
\end{tabularx}

\caption{Gender accuracy of Llama 2 70B Chat model and ChatGPT 3.5 model on the GATE test set. Gray text denote baseline results without the gender specified.}
\label{tab:gate_accuracy}
\vspace{-5mm}
\end{table}

Gender accuracy results of the various subsets of the GATE test set are reported in Table~\ref{tab:gate_accuracy}.
First, we find that the baseline translations of LLMs without gender control default to masculine inflections by a ratio of approximately 9:1. 
With GoE prompting, we observe over 95\% accuracy on male entities for both Llama 2 and ChatGPT. However, the accuracy of female entities is lower in comparison, indicating room for improvement of controlled translation with LLMs.

Also, we generally find that the gender accuracy of sentences containing multiple ambiguous entities (Acc\textsubscript{$\geq$2}) is slightly lower than those containing a single ambiguous entity (Acc\textsubscript{1}).
This trend potentially suggests that LLMs find controlling gender inflections of multiple entities within a sentence more challenging. In order to investigate this further, we take the GATE subset with exactly two ambiguous entities and compare the gender accuracy of samples where the two genders are assigned the same gender with the samples where they are assigned differently.

\begin{figure}[t]
\centering{
\includegraphics[width=\columnwidth]{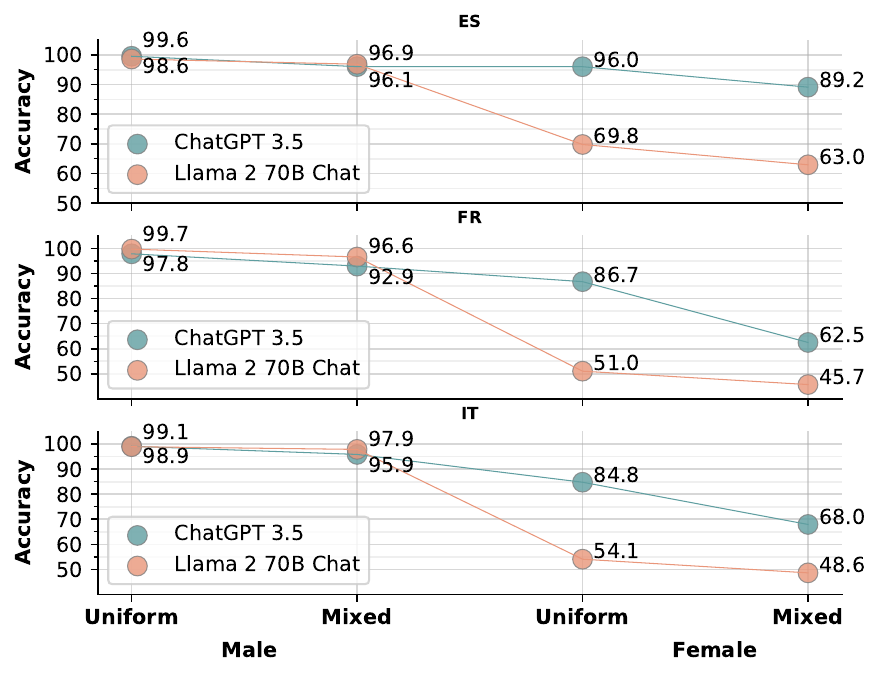}
}
\vspace{-5mm}
\caption{Gender accuracy of GoE prompting on the GATE subset with two ambiguous entities (\#Ent=2). \textit{Uniform} denotes translation with both entities mapped to the same gender, and \textit{Mixed} denotes translation with entities mapped to different genders.}
\vspace{-5mm}
\label{fig:gate_accuracy_by_subset}
\end{figure}

We report our fine-grained analysis of sentences with two ambiguous entities in Figure~\ref{fig:gate_accuracy_by_subset}.
Results demonstrate that LLMs find it easier to translate sentences with the same gender inflection for all entities (\textit{Uniform}) than those with mixed gender inflections (\textit{Mixed}). This \textit{interference} behavior in mixed settings adversely affects female gender mappings more strongly, with an absolute accuracy difference of up to 24.2\% between uniform and mixed settings.

\subsection{Gender Control of Mixed Entities}

In this subsection, we evaluate on the controlled translation of sentences containing a mix of ambiguously gendered and unambiguously gendered entities via the WinoMT benchmark \citep{stanovsky-etal-2019-evaluating}. 
We evaluate controlled translation with the gender of the ambiguous entity specified to have a \textit{different gender} from the existing unambiguous entity, as we observed in previous subsection~\ref{Gender Control of Multiple Ambiguous Entities} that having same genders for the multiple entities can artificially boost the performance.

\subsubsection{Baseline methods}

We compare our approach with a recent gender debiasing approach based on gender-aware contrastive learning (GACL) \citep{lee2023targetagnostic}. 
While their approach was originally proposed for improving the gender accuracy of unambiguously gendered entities, we also evaluate its effect on the ambiguously gendered entitis.
We evaluate the model based on NLLB-200 1.3B distilled model.

\subsubsection{Results}

\begin{table}[t]
    \centering
    \small
\setlength\tabcolsep{4.5pt}
\begin{tabularx}{\columnwidth}{X|rr|rr|rr}
\toprule
 & \multicolumn{2}{c}{\textbf{ES}} & \multicolumn{2}{c}{\textbf{FR}} & \multicolumn{2}{c}{\textbf{IT}} \\
 
\textbf{Method} & Acc\textsubscript{U} & Acc\textsubscript{A} &  Acc\textsubscript{U} & Acc\textsubscript{A} &  Acc\textsubscript{U} & Acc\textsubscript{A} \\
\midrule
\textit{NLLB-200} & & & & & & \\
Baseline & 72.0 & \color{gray}34.0 & 66.7 & \color{gray}36.6 & 54.1 & \color{gray}34.6 \\
GACL & \textbf{91.2} & \color{gray}3.0 & \textbf{85.0} & \color{gray}7.8 & \textbf{72.3} & \color{gray}7.3 \\
\midrule
\textit{Llama 2} & & & & & & \\
Baseline & 56.7 & \color{gray} 43.6 & 54.6 & \color{gray}44.2 & 46.5 & \color{gray} 41.8 \\
GoE\textsubscript{amb.} & 44.1 & 92.4 & 45.8 & \textbf{89.9} & 37.7 & \textbf{85.7} \\
GoE\textsubscript{full} & 74.3 & 83.7 & 67.0 & 69.4 & 65.0 & 75.7 \\
\midrule
\textit{ChatGPT} & & & & & & \\
Baseline & 62.4 & \color{gray}41.6 & 58.0 & \color{gray}41.8 & 49.6 & \color{gray}38.5 \\
GoE\textsubscript{amb.} & 39.2 & \textbf{93.9} & 42.7 & 84.5 & 37.0 & 76.8 \\
GoE\textsubscript{full} & 84.3 & 91.1 & 76.9 & 82.7 & 63.2 & 74.5 \\
\bottomrule
\end{tabularx}
\caption{Results on the WinoMT dataset. Acc\textsubscript{U} denotes the gender accuracy of the unambiguously gendered entity and Acc\textsubscript{A} denotes the gender accuracy of the ambiguously gendered entity.}
\vspace{-5mm}
\label{tab:winomt}
\end{table}

\begin{table*}[t]
    \centering
    \scriptsize
\setlength\tabcolsep{4pt}
\begin{tabularx}{\textwidth}{X|ccccc|ccccc|ccccc}
\toprule
 & \multicolumn{5}{c}{\textbf{ES}} & \multicolumn{5}{c}{\textbf{FR}} & \multicolumn{5}{c}{\textbf{IT}} \\
 
\textbf{Method} & Cov. & Acc\textsubscript{M} & Acc\textsubscript{F} & \textsc{bleu} & \textsc{comet} & Cov. & Acc\textsubscript{M} & Acc\textsubscript{F} & \textsc{bleu} & \textsc{comet}  & Cov. & Acc\textsubscript{M} & Acc\textsubscript{F} & \textsc{bleu} & \textsc{comet} \\
\midrule
\textit{NLLB-200 1.3B D.} & & & & & & & & & & & & & & & \\
Baseline                                          & 71.7 & \textbf{99.3} & 70.0 & 44.3 & 86.5 & 58.7 & 97.8 & 54.0 & 36.2 & 84.0 & 56.8 & 98.6 & 52.0 & 29.0 & 84.4 \\
GACL \scriptsize \citep{lee2023targetagnostic}    & 72.1 & 99.1 & \textbf{97.0} & 39.8 & 85.6 & 58.8 & \textbf{98.4} & 87.5 & 33.8 & 83.6 & 56.7 & \textbf{99.3} & \textbf{91.4} & 22.7 & 82.5 \\
\midrule
\textit{Llama 2 70B Chat} & & &  & & & & & & & & & & & & \\
Baseline        & 66.5 & 98.8 & 67.9 & 43.7 & 86.7 & 56.0 & 97.0 & 55.7  & 35.3 & 84.4 & 55.4 & 98.6 & 49.5 & 29.0 & 86.1 \\
GoE             & 67.8 & 99.0 & 78.6 & 44.1 & 87.1 & 58.2 & 97.3 & 70.0 & 37.1 & 85.3 & 54.9 & \textbf{99.3} & 62.0 & 29.6 & 86.5 \\
\midrule
\textit{ChatGPT 3.5} & & & & & & & & & & & & & & & \\
Baseline        & 71.6 & 97.1 & 86.2 & 48.1 & \textbf{89.3} & 61.7 & 97.6 & 81.4 & 41.8 & 87.8 & 59.0 & 98.7 & 68.7 & 33.5 & 88.7 \\
GoE             & 71.6 & 98.7 & 94.4 & 48.8 & 88.8 & 61.2 & 97.1 & \textbf{89.7} & 41.6 & 87.6 & 58.9 & 98.9 & 81.2 & 33.6 & 88.9 \\
Baseline (few-shot) & 72.3 & 98.4  & 85.8 & \textbf{50.9} & 89.7 & 66.5 & 97.1 & 86.5 & 43.6 & \textbf{88.2} & 61.7 & 98.3 & 72.7 & 34.8 & 89.0 \\
I-GoE (few-shot) & 71.8 & 98.6 & 90.4 & 50.5 & \textbf{89.3} & 61.9 & 98.0 & 88.0 & \textbf{43.8} & 87.8 & 59.4 & 98.5 & 76.3 & \textbf{35.0} & \textbf{89.0} \\
\bottomrule
\end{tabularx}
\caption{Results of controlled translation on the MT-GenEval Contextual test set.}
\vspace{-5mm}
\label{tab:mtgeneval}
\end{table*}

From the results shown in Table~\ref{tab:winomt}, we first notice that almost all baseline models score over 50\% on the gender accuracy of unambiguous entity (Acc\textsubscript{U}) while scoring lower than 50\% on the ambiguous entity (Acc\textsubscript{A}).
This indicates that instead of defaulting to masculine translation for ambiguous gender, models inflect it to the same gender as the unambiguous entity. 
This gender interference is amplified by the GACL method, where Acc\textsubscript{A} simultaneously drops by over 20\% as Acc\textsubscript{U} improves by 20\% compared to baseline.

Next, we find that explicitly controlling the gender of the ambiguous entity with GoE (GoE\textsubscript{amb.}) significantly improves the accuracy of Acc\textsubscript{A} for both Llama 2 and ChatGPT 3.5 models.
However, this time, we observe that Acc\textsubscript{U} is lower by at least 10\% compared to the baseline. 
These findings denote that both fine-tuning and GoE prompting methods interfere with the gender of other entities in the sentence. 
Also, on manual inspection, we find that a few of the WinoMT evaluation samples are inherently ambiguously phrased so that either of the entities could be referred by the gendered pronoun. 

Lastly, we experiment with controlling the gender of both entities with gold annotations using GoE prompting (GoE\textsubscript{full}).
Results show that explicitly specifying both entities leads to the best balanced accuracy improvement of both entities for both LLMs. 
These results suggest the usefulness of controlled translation for facilitating the correct translation of unambiguous entities, even if it can be inferred via coreference resolution. 

\subsection{Gender Control of Complex Unambiguous Entities}

In our fourth evaluation task, we evaluate controlling the gender of complex unambiguous entities using the \textit{Contextual} subset of the MT-GenEval dataset \citep{currey-etal-2022-mt}.
We experiment controlled translation by specifying the gender of the unambiguous entity from the second sentence in our prompt.
However, the dataset does not provide annotations on the unambiguous entity nor its gender. 
We thus obtain pseudo-gold entity annotation by using the Spacy\footnote{\url{https://spacy.io}} dependency parser to extract the noun phrase of the second sentence while using the gendered word list \citep{zhao-etal-2017-men} to extract the gender of the entity in the first sentence.

\subsubsection{Results}

In the results shown in Table~\ref{tab:mtgeneval}, we first note that baseline models show a relatively high gender accuracy compared to other evaluation datasets, as the evaluated entities are unambiguous and their gender can be inferred from the first sentence. 
Next, we find that explicitly specifying the pseudo-gold gender via GoE prompting improves the gender accuracy further, especially for the female gender with an improvement of 12.4\% and 9.5\% across evaluated language directions for Llama 2 and ChatGPT respectively. Translation quality remains within similar range before and after prompting, suggesting gender prompting does not harm the translation quality.

\subsubsection{Additional results on end-to-end translation}

Unlike ambiguously gendered entities, the gender of unambiguous entities can be inferred from the given text. 
Thus, we additionally experiment whether LLMs could be instructed to infer the entity and its gender from the given sentence and subsequently translate the sentence, in an end-to-end setup.
This idea is adopted from recent findings that generating intermediate reasoning steps improve performance of LLMs on complex reasoning tasks, since identifying the gender of entities could be seen as an intermediate reasoning step to generating translation with correct gender inflection. 

To instruct LLMs to \textbf{I}nfer the entity's gender and subsequently translate, we additionally add few-shot examples to the GoE prompt, which we refer to as \textbf{I-GoE prompting}. The few-shot examples are sampled from the MT-GenEval dev set, and the output translations start with the following pretext: ``\textit{From the given source text, we can infer that \texttt{[ENT]} uses \texttt{[GENDER]}. Therefore, the \texttt{[LANG]} translation with correct gender inflection is:}''.

Results of I-GoE prompting on ChatGPT shown in Table~\ref{tab:mtgeneval} show a meaningful improvement from the baseline, with an average of 5.8\% absolute improvement in female gender accuracy. 
However, the original GoE prompting based on pseudo-gold annotations still hold the highest gender accuracy overall, suggesting rooms for improvement in I-GoE prompting.

\subsection{Summary of Controlled Translation Experiments}

In this section, we evaluated the capability of LLMs to control gender inflections in MT for four different scenarios. Results showed that LLMs are highly capable of controlling the gender inflection for a single entity, but shows degradation in performance for multiple entities, especially when they have non-uniform gender assignments. Finally, we found that explicitly stating the gender inflection helps improve accuracy for unambiguously gendered entities as well, and using a two-step gender extraction and translation pipeline via I-GoE prompting moderately improves gender accuracy of the model. 

\section{LLM as Gender Evaluators}
In Section~\ref{Main Experiments}, our methodology exhibits significant performance based on automated gender accuracy metrics. However, the employed coverage-based metrics are dependent on the annotated gender terms. Such dependence poses a challenge in assessing gender terms that do not match the annotations due to the use of synonyms or different grammatical structures. For example, the English term ``professor'' can be translated into either "profesor/profesora" or "maestro/maestra" in Spanish. As shown in Tables~\ref{tab:mustshe}, \ref{tab:gate_accuracy}, and \ref{tab:mtgeneval}, at least 20\% of the samples for each evaluation benchmark remain unevaluated due to missing coverage from the provided gender annotations. As a result, it is necessary to address these issues for a more complete and accurate assessment of gender accuracy.

To address such complexity, we propose LLMs as Gender Evaluators (LGE). 
We provide LLMs with instructions as specified in Table~\ref{tab:prompt_eval} in Appendix~\ref{sec:appendix_eval}. As input, the LLM is given the source sentence, the model prediction, the controlled entity and its designated gender in English. It is then prompted to evaluate whether the given entity is inflected to the designated gender as a binary judgement of either \texttt{ACCURATE} or \texttt{INACCURATE}. Unlike existing coverage-based metric, our evaluation method does not require the reference translation nor any gendered term annotations in the target language, allowing evaluation of samples previously skipped due to limited coverage.

We explore the viability of LGE by first performing a sanity check with evaluation of gold human-provided translations. Then, we collect human expert annotations to assess the correlation of LGE with human judgements. Finally, we re-assess controlled translation with LGE, including samples previously omitted by the coverage-based metrics.

\subsection{Sanity Check with Reference Translations}
\label{subsec:Sanity Check with Reference Translations}
\begin{table}[]
\centering
\small
\begin{tabular}{@{}l|c|ccc@{}}
\toprule
\textbf{Dataset} & \textbf{Lang.} & \textbf{F1-score} & \textbf{Precision} & \textbf{Recall} \\
\midrule
\multirow{3}{*}{Must-SHE} & ES & 95.6 & 94.8 & 96.3 \\
& FR & 93.2 & 89.7 & 97.0 \\
& IT & 96.0 & 94.8 & 97.3 \\
\midrule
\multirow{3}{*}{GATE} & ES & 95.5 & 93.3 & 97.9 \\
& FR & 86.5 & 81.4 & 92.3 \\
& IT & 92.0 & 87.1 & 97.5 \\
\bottomrule
\end{tabular}
\caption{Sanity check results of LGE gender accuracy evaluation on Must-SHE and GATE test sets.}
\vspace{-2mm}
\label{tab:sanitycheck}
\end{table}

Initially, we conduct a sanity check to determine whether LLMs possess the capability to function as gender accuracy evaluators. The Must-SHE and the GATE dataset provide a valuable resource for this purpose, as they contain the possible variants of translations based on the gender of entities in the source sentences. Therefore, we conduct an experiment using these reference translations. In scenarios where the provided reference aligns with the specified gender condition, the LLMs should evaluate it as \texttt{ACCURATE}. Such correct references are considered positive samples. On the other hand, in cases where the provided reference is incorrect, the response should be \texttt{INACCURATE}, and these incorrect references are categorized as negative samples. We calculate the F1 score, precision, and recall based on this categorization. The results of these experiments are presented in Table~\ref{tab:sanitycheck}. When correct references are provided, the LLMs predominantly evaluate them as accurate. Conversely, when incorrect references are given, the models mostly evaluate them inaccurate. This results show the effectiveness of LLMs as gender accuracy evaluators. Experimental details are in Appendix~\ref{sec:appendix_eval}.

\subsection{Gender Accuracy Evaluation with LGE}
\label{subsec:Comparison with Human Evaluation}
\begin{table}[]
 \centering
 \small
\begin{tabular}{@{}l|c|c@{}}
\toprule
& Agreement ($\%$) & Cohen's $\kappa$     \\ \midrule
LGE \(\Leftrightarrow\) Human      & 93.0        & 0.691 \\
LGE \(\Leftrightarrow\) Cov.-based*  & 87.0        & 0.688 \\ \bottomrule
\end{tabular}
\caption{Agreement and Cohen's Kappa Coefficient between LGE, human labels, and the coverage-based metric. *Comparison between LGE and coverage-based metric is done with the subset covered by the coverage-based metric.}
\vspace{-5mm}
\label{tab:human_comparison}

\end{table}

Subsequently, we assess the validity of our LGE utilizing outputs of ChatGPT and NLLB-200 models on the MT-GenEval dataset. We sample 100 English-Spanish outputs covered by annotated gender terms and another 100 outputs not covered and thus unevaluable by existing metric. These are then compared with evaluations from human annotators and those based on gender terms. For outputs not covered by gender terms, we rely exclusively on human annotator evaluations. Details of the human annotation are in Appendix \ref{subsec:appendix_human_anno}. In Table~\ref{tab:human_comparison}, we observe a substantial agreement between LGE evaluations and human evaluations. This indicates the feasibility of using LGE to effectively evaluate outputs, regardless of whether they are covered by gender terms or not. Also, in cases covered by reference gender terms, there is a high correlation between coverage-based accuracy metric and LGE.

\begin{table}[]
\resizebox{\columnwidth}{!}{%
\begin{tabular}{@{}c|l|c|c@{}}
\toprule
\multicolumn{1}{l}{} & \multicolumn{1}{l|}{} & \multicolumn{2}{c}{\textbf{Translator Model}} \\      
\multicolumn{2}{c|}{}                            & \multicolumn{1}{l|}{ChatGPT + GoE} & \multicolumn{1}{l}{Llama 2 + GoE} \\
\midrule
\multicolumn{2}{c|}{Coverage}        &  67.0                                 &               62.2                    \\
\midrule
                              \multirow{2}{*}{Cov.-based}     & Acc\textsubscript{C}    & 96.6                              & 84.9                              \\
                                  & Acc\textsubscript{N.C}& \texttt{N/A}                 & \texttt{N/A}                                 \\
\midrule
\multirow{3}{*}{\textbf{LGE}}    & Acc\textsubscript{C}     & 94.7                             & 82.6                            \\
                                  & Acc\textsubscript{N.C} & 79.9                             & 64.8                             \\
                                  & Acc\textsubscript{All}        & 90.6                             & 76.6  \\
\bottomrule
\end{tabular}%
}
\caption{Re-evaluation results of our gender-controlled translation with LGE on the GATE dataset. Acc\textsubscript{C} represents gender accuracy on sentences covered by reference gender terms, and Acc\textsubscript{N.C} represents gender accuracy on sentences not covered.}
\vspace{-5mm}
\label{tab:resultwithLLM}
\end{table}

After ensuring the reliability for LGE, we re-examine the performance of our GoE prompting method with our new evaluation method. Evaluation results are shown in Table~\ref{tab:resultwithLLM}. In situations where the translation output is covered by the reference gender terms, LGE evaluation shows a level of accuracy similar to that of coverage-based metric. However, for sentences not covered by the reference gender terms, a tendency towards lower gender accuracy is observed. Our evaluation, being reference-free, allows us to uncover such situations. In cases of non-coverage, there is a higher likelihood that gender translation has not been accurately rendered. Therefore, metrics calculated only in cases of coverage should be interpreted as relative comparisons and not absolute values, as they might slightly overestimate the actual performance.

\section{Related Works}
\subsection{Ambiguous Gender in Machine Translation}
The problem of handling ambiguous gender in machine translation has been pointed out by multiple studies, providing benchmarks for evaluation \citep{Cho2019OnMG, Bentivogli2020GenderID, Rarrick2023GATEAC}.

Multiple approaches have been proposed to handle ambiguous gender bias in machine translation, including rewriting a translation to another gender \citep{Rarrick2023GATEAC}, generating gender-neutral translations \citep{Piergentili2023FromIL,Piergentili2023HiGO}, and controlled translation \citep{Bentivogli2020GenderID, sarti-etal-2023-ramp}. However, they do not consider fine-grained gender control of multiple entities.

A recent work also proposed gender-specific machine translation with LLMs \citep{Sanchez2023GenderspecificMT}. However, they also only consider two gendered variations for each sentence, and use LLMs to translate both variations without control.

\subsection{Machine Translation with LLMs}

As LLMs are widely adopted to various fields, recent studies have explored usage of LLMs for machine translation \cite{herold-etal-2023-improving,garcia2023unreasonable}. Despite being trained mainly on English corpora and with only limited number of parallel text, LLMs have shown competitive performance in machine translation without additional fine-tuning \citep{ vilar-etal-2023-prompting}. Additionally, the adoption of LLMs in MT has been shown to contribute to addressing diverse gender biases including pronoun genders and name entities \cite{Saunders2023GenderNA, wang-etal-2022-measuring, petrick-etal-2023-document, zhang2023prompting, attanasio2023tale}.

\subsection{LLM-based evaluation}
Traditionally, semantic-based metrics employ neural networks through encoder models such as BERTScore \cite{zhang2020bertscore}. Recently LLM-Eval \citep{lin-chen-2023-llm} utilized decoder-based models as metrics, and demonstrates a higher correlation with human evaluation. In MT tasks, \citet{kocmi-federmann-2023-large} shows GPT evaluation is better than BLEU.

\section{Conclusion}
In this paper, we tackled fine-grained gender control in machine translation. 
To solve this task, we proposed Gender-of-Entity prompting method for LLMs, where we instruct LLMs to translate with additional entity-level gender information given in natural language statements. 
Results on four evaluation benchmark show promising capabilities of LLMs as controlled translator of gender, with up to 95\% average accuracy on the MuST-SHE dataset.
We also observe a new phenomena of performance degradation when translating sentences with multiple gendered sentences with different target genders, which we refer to as gender interference.
Finally, we addressed the limitations of existing automated gender evaluation metrics by proposing LLMs as Gender Evaluators (LGE). 
Based on experimental results, LGE evaluations were shown to have high correlation with human judgements.

\section{Limitations}
Our study evaluates controlled translation in three languages that are supported by all four evaluation benchmarks, Spanish, French, and Italian, to allow multi-faceted analysis and comparison.
The three languages all fall within the Romance language family and often categorized as a high resource language. 
Hence, further investigation is required on low-resource languages and other languages not covered by our study for evaluating the controlled translation performance of LLM.

Additionally, the utilization of GoE prompting and its evaluation requires the gender-annotated dataset. Particularly, if the annotations contain errors, there is a possibility that it could actually lead to a degradation in performance. 
To address such problem in our research, we make 
evaluation methods extending the setting of existing studies. However, given the inherent complexity and intricate nature of languages, there may still be instances where our approach fails to adequately address scenarios involving sentences that lack explicit entities or where both ambiguous and unambiguous entities are intricately intertwined.

Finally, even though our methodology demonstrates great performance compared to baselines, there is much room for improvement. 
Some possible future directions include improving translation with few-shot examples, constructing a more sophisticated instruction prompt, and incorporating reinforcement learning.

\section{Ethical Considerations}
Since our research is concentrated on gender bias related to ambiguous entities, the applicability of our study to other demographic biases beyond gender remains under-explored. Therefore, any extension of our methodology to encompass demographic biases would require thorough consideration and additional research.

Furthermore, since annotations in existing datasets are framed within a binary setting, we have limited results only on the binary gender, difficult to evaluate performance on gender-netural or non-binary genders in our studies. 
However, as Multilingual Large Language Models (LLMs) have shown to well-adapt to tasks with instructions, we believe that, given the availability of relevant datasets, our methodology could also be applicable to non-binary genders. 

\section*{Acknowledgements}

This work was supported by Institute of Information \& communications Technology Planning \& Evaluation (IITP) grant funded by the Korea government(MSIT) [NO.2022-0-00184, Development and Study of AI Technologies to Inexpensively Conform to Evolving Policy on Ethics]. 
This work was partly supported by Institute of Information \& communications Technology Planning \& Evaluation (IITP) grant funded by the Korea government(MSIT) [NO.2021-0-01343, Artificial Intelligence Graduate School Program (Seoul National University)].
K. Jung is with Automation and Systems Research Institute (ASRI), Seoul National University.

\bibliography{anthology,ref}

\clearpage
\newpage
\appendix

\section{Experimental Details for Controlled Translation}
\label{sec:appendix}

\subsection{Dataset Statistics}

We report the dataset statistics of the four evaluation benchmarks used in this paper in Table~\ref{tab:data_stat}. 
For the GATE dataset, we evaluate all entity-level gender mapping combinations for each sample. Hence, the number of evaluated translations is equal to the number of dataset samples multiplied by the number of possible mappings, which is $2^N$, where $N$ is the number of ambiguous entities.

For the GATE dataset, we exclude samples with incorrect annotations, where the number of entities does not match the annotations. 
For the MT-GenEval dataset, we exclude samples with incorrect annotations, where the first sentence is either blank or does not contain gendered terms based on the word list. We manually went over the excluded samples to verify that the annotation was incorrect. 
For computing the term-based accuracy, we use the gold annotated gender terms and entity terms for MuST-SHE and GATE datasets. For the MT-GenEval dataset, we obtain gendered terms by comparing and extracting the differing terms between the male and female gold reference translations provided by the dataset. For the diff tool, we use the \texttt{difflib.SequenceMatcher} algorithm in the Python 3 -build-in library.

\subsection{LLM output post-processing}

In our translation experiments with LLMs, we found LLMs often generate additional comments either before or after the translations. Thus, we apply a basic rule-based post-processing to extract the translated sentence from the LLM generation output. First, we split the output into sentences based on the newline character \texttt{\textbackslash n}, and filter out sentences that contain any of the following tokens: ``gender'', ``translat'', ``sentence'', and ``note''. Out of the remaining sentences, we take the first sentence as the translation output.


\begin{table}[h]
    \centering
    \small
\begin{tabularx}{\columnwidth}{Xcrrr}
\toprule
\textbf{Dataset} & \textbf{Subset} & \textbf{ES} & \textbf{FR} & \textbf{IT} \\
\midrule
\multirow{2}{*}{MuST-SHE} & 1M & 287 & 292& 282 \\
                          & 1F & 284 & 315& 278 \\
\midrule
\multirow{3}{*}{GATE test} & \#Ent=1 & 751 & 775 & 564 \\
                      & \#Ent=2 & 150 & 222 & 259 \\
                      & \#Ent=3 & 12  & 0   & 20  \\
\midrule
\multirow{1}{*}{WinoMT} & & 3,888 & 3,888 & 3,888 \\
\midrule
\multirow{1}{*}{MT-GenEval test} & Contextual & 1,096 & 1,099 & 1,094 \\

\bottomrule
\end{tabularx}
\caption{Dataset statistics of the four evaluated benchmarks.}
\label{tab:data_stat}
\end{table}

\section{Experimental details for LLMs as Gender Evaluators}
\label{sec:appendix_eval}

\begin{table}[]
\resizebox{\columnwidth}{!}{
\begin{tabular}{@{}l|c|ccc|ccc@{}}
\toprule
 && \multicolumn{3}{c}{gpt-4-turbo} & \multicolumn{3}{c}{gpt-3.5-turbo} \\
\textbf{Dataset} &Lang.& F1-score & Precision & Recall & F1-score & Precision & Recall \\
\midrule
Must-SHE & ES &\textbf{95.6} & 94.8 & 96.3 & 71.8 & 56.4 & 98.8 \\
GATE & ES& \textbf{95.5} & 93.3 & 97.9 & 57.4 & 40.4 & 98.8\\
\bottomrule
\end{tabular}
}
\caption{Sanity Check Results for ChatGPT models.}
\label{tab:comparison_sanity}
\end{table}

\begin{table}[h]
\resizebox{\columnwidth}{!}{
\begin{tabular}{l|c|c|cc}
\toprule
&  & & \multicolumn{2}{c}{\textbf{Gender Accuracy}} \\
\textbf{Dataset} & \textbf{\#Ent} & \textbf{Lang.} & w/ Correct Ref. & w/ Wrong Ref. \\ 
\midrule
{\multirow{8}{*}{GATE}} & {\multirow{3}{*}{1}} & ES & 98.7 & 1.1 \\
 &  & FR & 92.6 & 8.7 \\
 &  & IT & 97.3 & 2.0 \\ \cmidrule{2-5} 
 & {\multirow{3}{*}{2}} & ES & 97.8 & 4.3 \\
 &  & FR & 91.7 & 14.4 \\
{} &  & IT & 97.7 & 6.4 \\ \cmidrule{2-5} 
 & {\multirow{2}{*}{3}} & ES & 85.3 & 9.8 \\
 &  & IT & 98.1 & 11.0 \\
\midrule
{\multirow{3}{*}{Must-SHE}} & {\multirow{3}{*}{1}} & ES & 96.3 & 5.3 \\
 &  & FR & 97.0 & 11.2 \\
 &  & IT & 97.3 & 5.5 \\
\bottomrule
\end{tabular}
}
\caption{Sanity check results for each reference sentence type and number of entities on GATE and Must-SHE datasets. Higher scores are better for evaluation with correct references, and lower scores are better for evaluation with wrong references.}
\label{tab:sanitycheck_reference}
\end{table}

\begin{table}[h]
    \centering
    \small
\begin{tabularx}{\columnwidth}{X}
\toprule
\textbf{LGE Evaluation Prompting Template} \\
\midrule
SYSTEM: \\ \\
        You are evaluating a gender-conditioned translation.
        Please specifically focus on whether the translation accurately reflects the gender representation of the provided entities.
        Check if the words related to the entities are translated in a way that is consistent with the entities' specified genders.
        After reviewing the input, provide your evaluation in the following format: \\ \\
        Comment: [Your explanation regarding the gender representation in relation to the entities in the translation.] \\ \\
        Gender Accuracy: [ACCURATE or INACCURATE].
\\ \\
USER: \\ \\ Source [EN]:  \texttt{[SRC]} \\
Condition: Entity "\texttt{[ENT]}" should be translated as "\texttt{[GEN]}"\\
Translation \texttt{[LANG]}: \texttt{[PRED]} \\
\bottomrule
\end{tabularx}
\caption{Instruction template for gender evaluation used by LGE.}
\label{tab:prompt_eval}
\end{table}

\begin{table}[h]
    \centering
    \small
\begin{tabularx}{\columnwidth}{X}
\toprule
\textbf{I-GoE Few-shot Prompting Template} \\
\midrule
SYSTEM: You are a professional \texttt{[TGT\_LANG]} translator that especially considers translating gender inflections correctly. \\ \\
USER: Help me translate the following source text into \texttt{[TGT\_LANG]}. \\
ASSISTANT: Sure, I’d be happy to! \\
USER: \texttt{[SRC\_1]}  \\
ASSISTANT: From the given source text, we can infer that \texttt{[ENT\_1]} uses \texttt{[GENDER\_1]}. Therefore, the \texttt{[TGT\_LANG]} translation with correct gender inflection is: \\
\texttt{[TGT\_1]} \\
USER: \texttt{[SRC\_2]} \\
ASSISTANT: From the given source text, we can infer that \texttt{[ENT\_2]} uses \texttt{[GENDER\_2]}. Therefore, the \texttt{[TGT\_LANG]} translation with correct gender inflection is: \\
\texttt{[TGT\_2]} \\
… \\
USER: \texttt{[SRC]} \\
ASSISTANT: \\
\bottomrule
\end{tabularx}
\caption{Instruction template for few-shot I-GoE prompting.}
\label{tab:prompt_igoe}
\end{table}

We utilize a state-of-the-art large language model,  gpt-4-turbo (\texttt{gpt-4-1106-preview}) for the role of gender accuracy evaluator. Our initial sanity check experiments revealed that gpt-3.5-turbo model showed less satisfactory performance for this role, as shown in Table \ref{tab:comparison_sanity}. 
The example prompts are in Table \ref{tab:prompt_eval}.

\subsection{Human Annotation Process}
\label{subsec:appendix_human_anno}
\begin{figure*}[t]  
\centering{
\includegraphics[width=\textwidth]{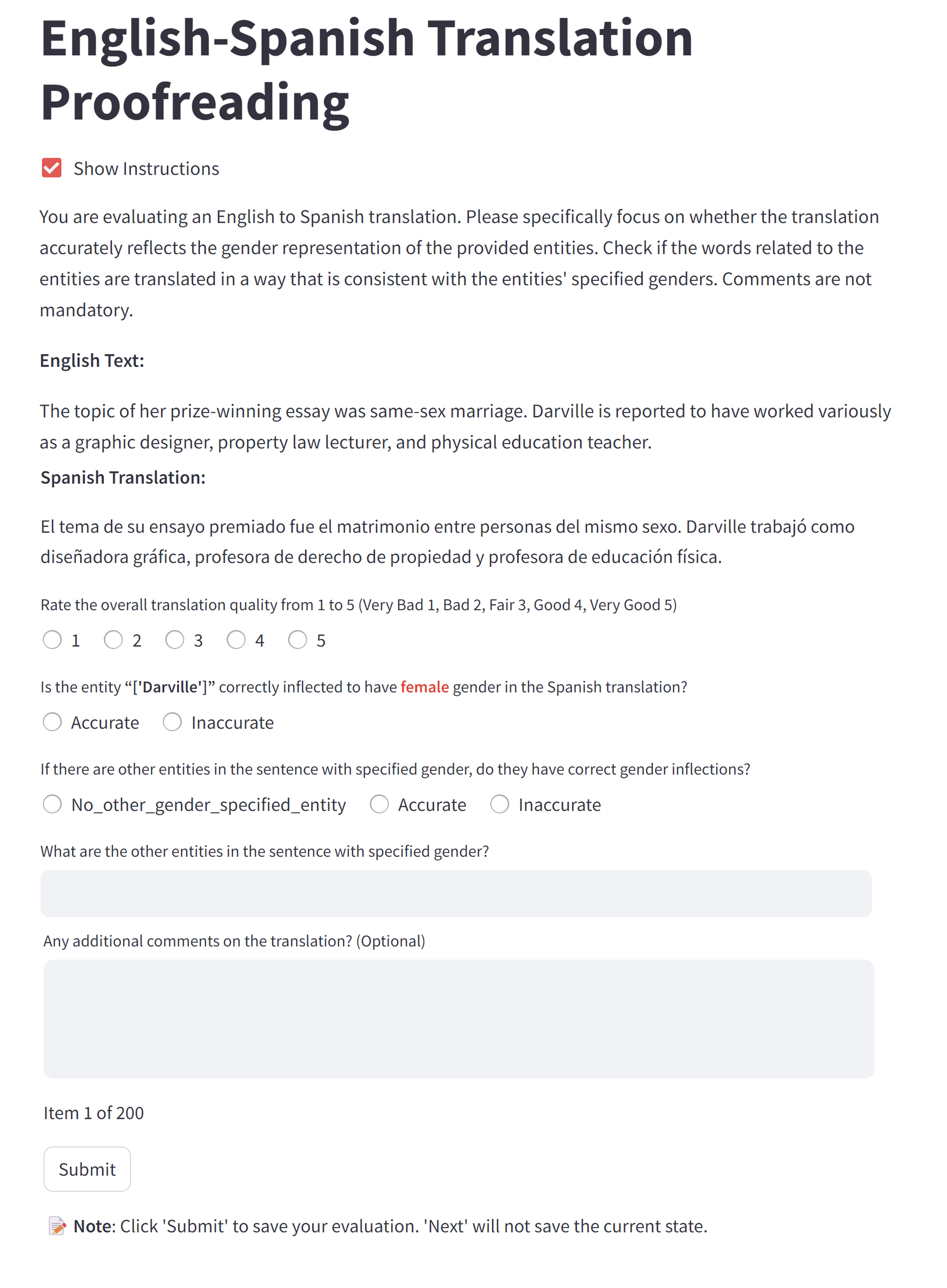}  
}
\caption{Example of human annotation pages}
\label{fig:human_page}
\end{figure*}

\begin{table}[]
 \centering
 \small
\begin{tabular}{@{}l|c|c@{}}
\toprule
& Agreement($\%$) & $\kappa$     \\ \midrule
Human \(\Leftrightarrow\) Ref-based*       & 84.0        & 0.607 \\ \bottomrule
\end{tabular}
\caption{Agreement and Cohen's Kappa Coefficient between the human annotators and the reference-based metric. Comparisons are calculated only for 100 outputs that are covered by reference gender terms.}
\label{tab:human_ref_comparison}
\end{table}

As described in Section \ref{subsec:Comparison with Human Evaluation}, we employ three Spanish-English bilingual annotators to evaluate the gender accuracy of ChatGPT and NLLB outputs based on the MT-GenEval dataset. 
Native Spanish speakers from the author's local communities proficient in both Spanish and English are recruited as annotators. They were informed with the research objective of this annotation and obtained consent on the use of the dataset.
We instruct the annotators to conduct a more detailed evaluation by assessing additional aspects beyond just gender accuracy. An example of the detailed annotation page can be found in Figure \ref{fig:human_page}. The final assessment of gender accuracy is achieved by adopting the majority decision among the three expert annotators. In the gender accuracy evaluation conducted by three annotators, there are 182 instances where all three agree, and 18 instances where only two agree. The agreement among these three, calculated using the Fleiss kappa coefficient, is 0.709, indicating that the opinions of the three annotators substantially agree. Additionally, the comparison between the final human label, determined by the majority voting of these three annotators, and the original reference-based metric, is presented in Table \ref{tab:human_ref_comparison}. Most of the discrepancies stem from errors by the annotators or label errors within the dataset itself.

\subsection{Example Outputs of LGE}
Example outputs of LGE can be found in Table \ref{tab:ex_out}.

\begin{table*}
\centering
\begin{tabularx}{\textwidth}{X}
\toprule
\texttt{User Prompt:} \\
Source [EN]: At that time, the Russian ambassador was not there. \\
Condition: Entity "ambassador" should be translated as feminine. \\
Translation [ES]: En ese momento, la embajadora rusa no estaba allí. \\\\

\texttt{Output:} \\
Comment: The translation accurately reflects the gender representation of the entity "ambassador" by using the feminine form "la embajadora" in Spanish, which aligns with the condition provided. \\
Gender Accuracy: ACCURATE \\
\midrule
\texttt{User Prompt:} \\
Source [EN]: This was an uneasy alliance at best, and the allies would periodically break off the assault to fight each other. \\
Condition: Entity "allies" should be translated as masculine. Entity "each other" should be translated as masculine. \\
Translation [ES]: Aquella fue una alianza frágil, en el mejor de los casos, pues los aliados interrumpieron varias veces el ataque para luchar entre ellas. \\\\
\texttt{Output:} \\
Comment: The translation of "allies" as "aliados" is correct and masculine, which is consistent with the specified gender. However, the phrase 'para luchar entre ellas' is incorrect because "ellas" is feminine and should be "ellos" to reflect the masculine gender specified for "each other". \\
Gender Accuracy: INACCURATE. \\
\midrule
\texttt{User Prompt:} \\
Source [EN]: In 1668, plays by women began to be staged in London. The performance ran for six nights, which was regarded as a good run for an unknown author. \\
Condition: Entity "author" should be translated as feminine. \\
Translation [ES]: En 1668, las obras de teatro de mujeres comenzaron a ser puestas en escena en Londres. La actuación duró seis noches, lo que se consideró una buena carrera para un autor desconocido. \\\\

\texttt{Output:} \\
Comment: The translation does not accurately reflect the gender of the entity "author" as feminine. The Spanish word for author, "autor" should have been translated as "autora" to indicate a female author. \\
Gender Accuracy: INACCURATE \\
\bottomrule
\end{tabularx}
\caption{Example Outputs of LGE}
\label{tab:ex_out}
\end{table*}

\end{document}